\documentclass[11pt,a4paper]{article}
\usepackage[hyperref]{emnlp-ijcnlp-2019}
\usepackage{times}
\usepackage{graphicx}
\usepackage{verbatim}
\usepackage{latexsym}
\usepackage{booktabs}
\usepackage{stfloats}
\usepackage{array}
\usepackage{bm}
\usepackage{amsfonts}
\usepackage{amsmath}
\usepackage{multirow}
\usepackage{bbding}

\usepackage{url}

\def\argmin{\mathop{\rm argmin}}%

\aclfinalcopy 

\title{Learning to Copy for Automatic Post-Editing}

\author{Xuancheng Huang$^\dag$, Yang Liu${^\dag}{^\ddag}{\thanks{\ \ Corresponding author: Yang Liu}}$, Huanbo Luan$^\dag$, Jingfang Xu$^\S$ and Maosong Sun$^\dag$
\\
  $^\dag$Institute for Artificial Intelligence \\
  State Key Laboratory of Intelligent Technology and Systems \\
  Department of Computer Science and Technology, Tsinghua University, Beijing, China\\
  Beijing National Research Center for Information Science and Technology\\
  $^\S$Sogou Inc., Beijing, China\\
  $^\ddag$Beijing Advanced Innovation Center for Language Resources\\
  {\tt \small hxc17@mails.tsinghua.edu.cn, liuyang2011@tsinghua.edu.cn, luanhuanbo@gmail.com,} \\
  {\tt \small xujingfang@sogou-inc.com, sms@tsinghua.edu.cn}
}

\date{}

\begin{document}
\maketitle
\begin{abstract}
Automatic post-editing (APE), which aims to correct errors in the output of machine translation systems in a post-processing step, is an important task in natural language processing. While recent work has achieved considerable performance gains by using neural networks, how to model the copying mechanism for APE remains a challenge. In this work, we propose a new method for modeling copying for APE. To better identify translation errors, our method learns the representations of source sentences and system outputs in an interactive way. These representations are used to explicitly indicate which words in the system outputs should be copied, which is useful to help CopyNet \cite{Gu2016IncorporatingCM} better generate post-edited translations. Experiments on the datasets of the WMT 2016-2017 APE shared tasks show that our approach outperforms all best published results. \footnote{The source code is available at \url{https://github.com/THUNLP-MT/L2Copy4APE}}
\end{abstract}

\section{Introduction}

Automatic post-editing (APE) is an important natural language processing (NLP) task that aims to automatically correct errors made by machine translation systems \cite{Knight1994AutomatedPO}. It can be considered as an efficient way to modify translations to a specific domain or to incorporate additional information into translations rather than translating from scratch \cite{McKeown2012CanAP, Chatterjee2015ExploringTP, Chatterjee2018FindingsOT}. 

Approaches to APE can be roughly divided into two broad categories: {\em statistical} and {\em neural} approaches. While early efforts focused on statistical approaches relying on manual feature engineering \cite{Simard2007StatisticalPP, Bchara2011StatisticalPF}, neural network based approaches capable of learning representations from data have shown remarkable superiority over their statistical counterparts \cite{Varis2017CUNISF, Chatterjee2017MultisourceNA, JunczysDowmunt2017AnEO,Unanue2018ASA}. Most of them cast APE as a multi-source sequence-to-sequence learning problem \cite{Zoph2016MultiSourceNT}: given a source sentence ($src$) and a machine translation ($mt$), APE outputs a post-edited translation ($pe$).

\begin{table}[!t]
    \begin{center}
        \begin{tabular}{|l|l|}
            \hline
            $src$ & I ate a cake yesterday \\
            \hline
            $mt$ & \textbf{Ich} esse \textbf{einen} Hamburger  \\
            \hline
            $pe$ & \textbf{Ich} hatte gestern \textbf{einen} Kuchen gegessen \\
            \hline
        \end{tabular}
        \caption{\label{tab:example} Example of automatic post-editing (APE). Given a source sentence ($src$) and a machine translation ($mt$), the goal of APE is to post-edit the erroneous translation to obtain a correct translation ($pe$). Our work aims to explicitly model how to copy words from $mt$ to $pe$ (highlighted in bold), which is a common phenomenon in APE.} \label{table:example}
    \end{center}
\end{table}

A common phenomenon in APE is that many words in $mt$ can be \textbf{copied} to $pe$. As shown in Table \ref{table:example}, two German words ``Ich'' and ``einen'' occur in both $mt$ and $pe$. Note that the positions of copied words in $mt$ and $pe$ are not necessarily identical (e.g., ``einen'' in Table \ref{table:example}). Our analysis on the datasets of the WMT 2016 and 2017 APE shared tasks shows that over 80\% of words in $mt$ are copied to $pe$. As APE models not only need to decide which words in $mt$ should be copied correctly, but also should place the copied words in appropriate positions in $pe$, it is challenging to model copying for APE. Our experiments show that the state-of-the-art APE method \cite{JunczysDowmunt2018MSUEdinST} only achieves a copying accuracy of 64.63\% (see Table \ref{tab:acc}). 


\begin{figure*}[!t]
  \centering
  \includegraphics[scale=0.7]{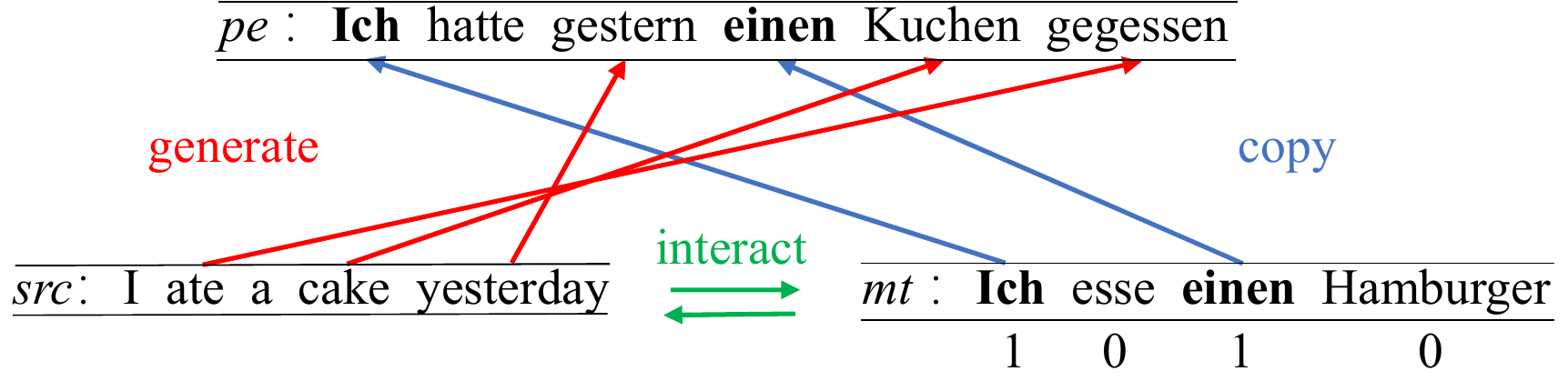}
  \caption{Learning to copy for APE. Our work is based on two key ideas. As both $src$ and $mt$ play important roles in APE, the first idea is that $src$ and $mt$ should ``interact'' with each other during representation learning to better generate words from $src$ and copy words from $mt$ during inference. The second idea is to predict which target words in $mt$ should be copied since it is easy to obtain labeled data automatically by comparing $mt$ and $pe$. The words to be copied (e.g., ``Ich'') in $mt$ are labeled with 1's and other words (e.g., ``esse'') with 0's.}
  \label{fig:example}
\end{figure*}

We believe that existing approaches to APE \cite{Varis2017CUNISF, Chatterjee2017MultisourceNA, JunczysDowmunt2017AnEO,Unanue2018ASA} suffer from two major drawbacks when modeling the copying mechanism. First, the representations of $src$ and $mt$ are learned separately. APE is a two-source sequence-to-sequence learning problem in which both $src$ and $mt$ play important roles. On the one hand, if $src$ is ignored, it is difficult to identify translation errors related to adequacy in $mt$, especially for fluent but inadequate translations (e.g., $mt$ in Figure \ref{fig:example}). On the other hand, $mt$ serves as a major source for generating $pe$ since many words (e.g., ``Ich" and ``einen" in Figure \ref{fig:example}) are copied from $mt$ to $pe$. Intuitively, it is likely to be easier to decide which words in $mt$ should be copied if $src$ and $mt$ fully ``interact'' with each other during representation learning.  Although CopyNet \cite{Gu2016IncorporatingCM} can be adapted for explicitly modeling the copying mechanism in multi-source sequence-to-sequence learning, the lack of the interaction between $src$ and $mt$ still remains a problem.

Second, there is no explicit labeling that indicates which target words in $mt$ should be copied. Existing approaches only rely on the attention between the encoder and decoder to implicitly choose target words to be copied. Given $mt$ and $pe$, it is easy to decide whether a target word in $mt$ should be copied or not. In Figure \ref{fig:example}, the words in $mt$ that should be copied are labeled with 1's. Other words are labeled with with 0's, which should be re-generated from $src$. These labels can served as useful supervision signals to help better copy words from $mt$ to $pe$, even when CopyNet is used. 

In this work, we propose a new method for modeling the copying mechanism for APE. As shown in Figure \ref{fig:example}, our work is based on two key ideas. First, our method is capable of learning the representations of input in an interactive way by enabling $src$ and $mt$ to attend to each other during representation learning. This might be useful for deciding when to generate words from $src$ and when to copy words from $mt$ during post-editing. Second, it is possible to predict which words in $mt$ should be copied because it is easy to automatically construct labeled data by comparing $mt$ and $pe$. Such predictions can be combined with CopyNet to better model copying for APE. Experiments show that our approach outperforms the best published results on the datasets of the WMT 2016-2017 APE shared tasks.

\section{Background}

\begin{figure*}[!t]
  \centering
  \includegraphics[width=\linewidth]{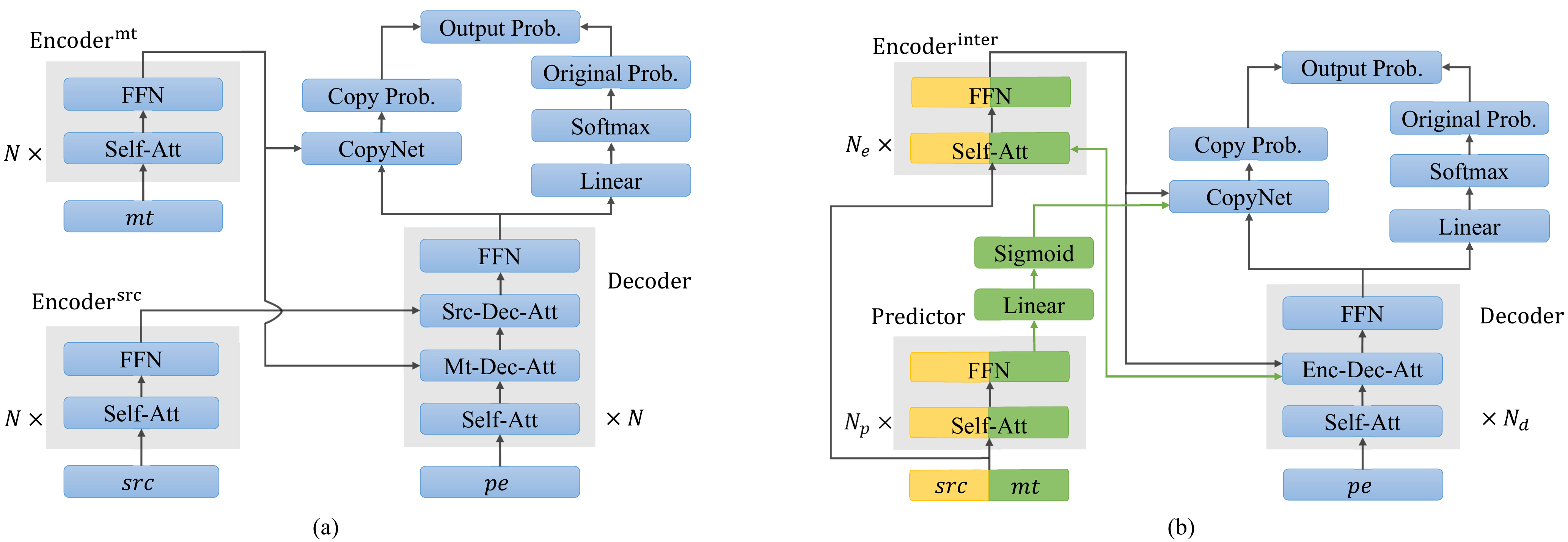}
  \caption{(a) The architecture of multi-source Transformer \cite{JunczysDowmunt2018MSUEdinST} equipped with CopyNet \cite{Gu2016IncorporatingCM} and (b) the architecture of our approach. While the existing work learns the representations of $src$ and $mt$ separately, our approach allows for learning the representations of $src$ and $mt$ in an interactive way by concatenating them as a single input. In addition, our approach introduces a Predictor module to explicitly indicate which words in $mt$ should be copied.}
  \label{fig:architecture}
\end{figure*}

\subsection{Multi-source Sequence-to-Sequence Learning}
Multi-source sequence-to-sequence learning has been widely used in APE in recent years \cite{JunczysDowmunt2018MSUEdinST, Pal2018ATM, Tebbifakhr2018MultisourceTF, Shin2018MultiencoderTN}. The architecture of multi-source Transformer is shown in Figure \ref{fig:architecture}(a). It can be equipped with CopyNet (see Section \ref{sec:CopyNet}) to serve as a baseline in our experiments. It is worth noting that $src$ and $mt$ are encoded separately.

Let $\mathbf{x}=x_1 \dots x_I$ be a source sentence (i.e., $src$) with $I$ words, $\mathbf{\tilde{y}}=\tilde{y}_1 \dots \tilde{y}_K$ be a translation output by a machine translation system (i.e., $mt$) with $K$ words, and $\mathbf{y}=y_1 \dots y_J$ be the post-edited translation (i.e., $pe$) with $J$ words. The APE model is given by
\begin{equation}
\begin{gathered}
P(\mathbf{y}|\mathbf{x}, \mathbf{\tilde{y}}; \bm{\theta}) = \prod_{j=1}^{J}P(y_j|\mathbf{x},\mathbf{\tilde{y}},\mathbf{y}_{<j}; \bm{\theta}), \label{eqn:overall_prob}
\end{gathered}
\end{equation}
where $y_j$ is the $j$-th target word in $pe$, $\mathbf{y}_{<j} = y_1 \dots y_{j-1}$ is a partial translation, $\bm{\theta}$ is a set of model parameters, and $P(y_j | \mathbf{x}, \mathbf{\tilde{y}}, \mathbf{y}_{<j}; \bm{\theta})$ is a word-level translation probability.

The word-level translation probability in Eq. (\ref{eqn:overall_prob}) is computed as 
\begin{alignat}{1}
\mathbf{H}^{\mathrm{src}} &= \mathrm{Encoder}^{\mathrm{src}}(\mathbf{x},\bm{\theta}), \\
\mathbf{H}^{\mathrm{mt}} &= \mathrm{Encoder}^{\mathrm{mt}}(\mathbf{\tilde{y}},\bm{\theta}), \\
\mathbf{h}^{\mathrm{pe}}_j &= \mathrm{Decoder}(\mathbf{y}_{<j}, \mathbf{H}^{\mathrm{src}}, \mathbf{H}^{\mathrm{mt}}, \bm{\theta}), \label{eqn:h_pe}\\
P(y_j|&\mathbf{x},\mathbf{\tilde{y}},\mathbf{y}_{<j}; \bm{\theta}) \propto \exp(\mathbf{h}^{\mathrm{pe}}_j\mathbf{W}_g),
\end{alignat}
where $\mathrm{Encoder}^{\mathrm{src}}(\cdot)$ is the encoder for $src$, $\mathbf{H}^{\mathrm{src}}$ is the real-valued representation of $src$, $\mathrm{Encoder}^{\mathrm{mt}}(\cdot)$ is the encoder for $mt$, $\mathbf{H}^{\mathrm{mt}}$ is the representation of $mt$, $\mathrm{Decoder}(\cdot)$ is the decoder, $\mathbf{h}^{\mathrm{pe}}_j$ is the representation of the $j$-th target word $y_j$ in $pe$. $\mathbf{W}_g \in \mathbb{R}^{d \times \mathcal{V}_y}$ is a weight matrix, $d$ is the dimension of hidden states, and $\mathcal{V}_y$ is the target vocabulary size.

A limitation of the aforementioned model is that $src$ and $mt$ are encoded separately without interacting with each other, which might lead to the inability to find which $src$ word is untranslated and which $mt$ word is incorrect. For example, the $mt$ sentence in Figure \ref{fig:example} is fluent and meaningful. Without $src$, the APE system is unable to identify translation errors. In addition, the multi-source Transformer does not explicitly model the copying between $mt$ and $pe$ in neither the Encoder nor the Decoder.

\subsection{CopyNet}
\label{sec:CopyNet}
CopyNet \cite{Gu2016IncorporatingCM} is a widely used method for modeling copying in sequence-to-sequence learning. It has been successfully applied to single-turn dialogue \cite{Gu2016IncorporatingCM}, text summarization \cite{See2017GetTT}, and grammar error correction \cite{Zhao2019ImprovingGE}.

It is possible to extend the multi-source Transformer with CopyNet to explicitly model the copying mechanism, as shown in Figure \ref{fig:architecture}(a). CopyNet defines the word-level translation probability in Eq. (\ref{eqn:overall_prob}) as a linear interpolation of copying and generating probabilities:
\begin{alignat}{1}
P(y_j|\mathbf{x},\mathbf{\tilde{y}},\mathbf{y}_{<j}; \bm{\theta}) = &\ \gamma_j \times P^{\mathrm{copy}}(y_j) \nonumber \\
 &+ (1-\gamma_j) \times P^{\mathrm{gen}}(y_j), \label{eqn:copy_gate}
\end{alignat}
where $P^{\mathrm{copy}}(y_j)$ is the copying probability for $y_j$, $P^{\mathrm{gen}}(y_j)$ is the generating probability for $y_j$, and $\gamma_j$ is a gating weight.  They are defined as follows:
\begin{alignat}{1}
P^{\mathrm{copy}}(y_j) \propto&\ \mathrm{exp}(g(\mathbf{H}^{\mathrm{mt}}, \mathbf{h}^{\mathrm{pe}}_j)), \label{eqn:p_copy}\\
P^{\mathrm{gen}}(y_j) \propto&\  \mathrm{exp}(\mathbf{h}^{\mathrm{pe}}_j\mathbf{W}_g), \\
\gamma_j =& \ u(\mathbf{H}^{\mathrm{mt}}, \mathbf{h}^{\mathrm{pe}}_j), \label{eqn:gamma}
\end{alignat}
where $g(\cdot)$ and $u(\cdot)$ are non-linear functions. See \cite{Zhao2019ImprovingGE} for more details.

Copying in APE involves two kinds of decisions: (1) choosing words in $mt$ to be copied and (2) placing the copied words in appropriate positions in $pe$. CopyNet makes the two kinds of decisions simultaneously. We conjecture that if which words in $mt$ should be copied can be explicitly indicated, it might be easier for CopyNet to copy words from $mt$ to $pe$ correctly. Therefore, it is necessary to design a new method for identifying words to be copied.


\section{Approach}

Figure \ref{fig:architecture}(b) shows the overall architecture of our approach. It differs from previous work in two aspects. First, we propose to let $src$ and $mt$ ``interact" with each other to learn better representations (Section \ref{sec:inter}). Second, our approach introduces a Predictor module to predict words to be copied (Section \ref{sec:pred}). Section \ref{sec:training} describes how to train our APE model.

\begin{figure}[!t]
  \centering
  \includegraphics[width=\linewidth]{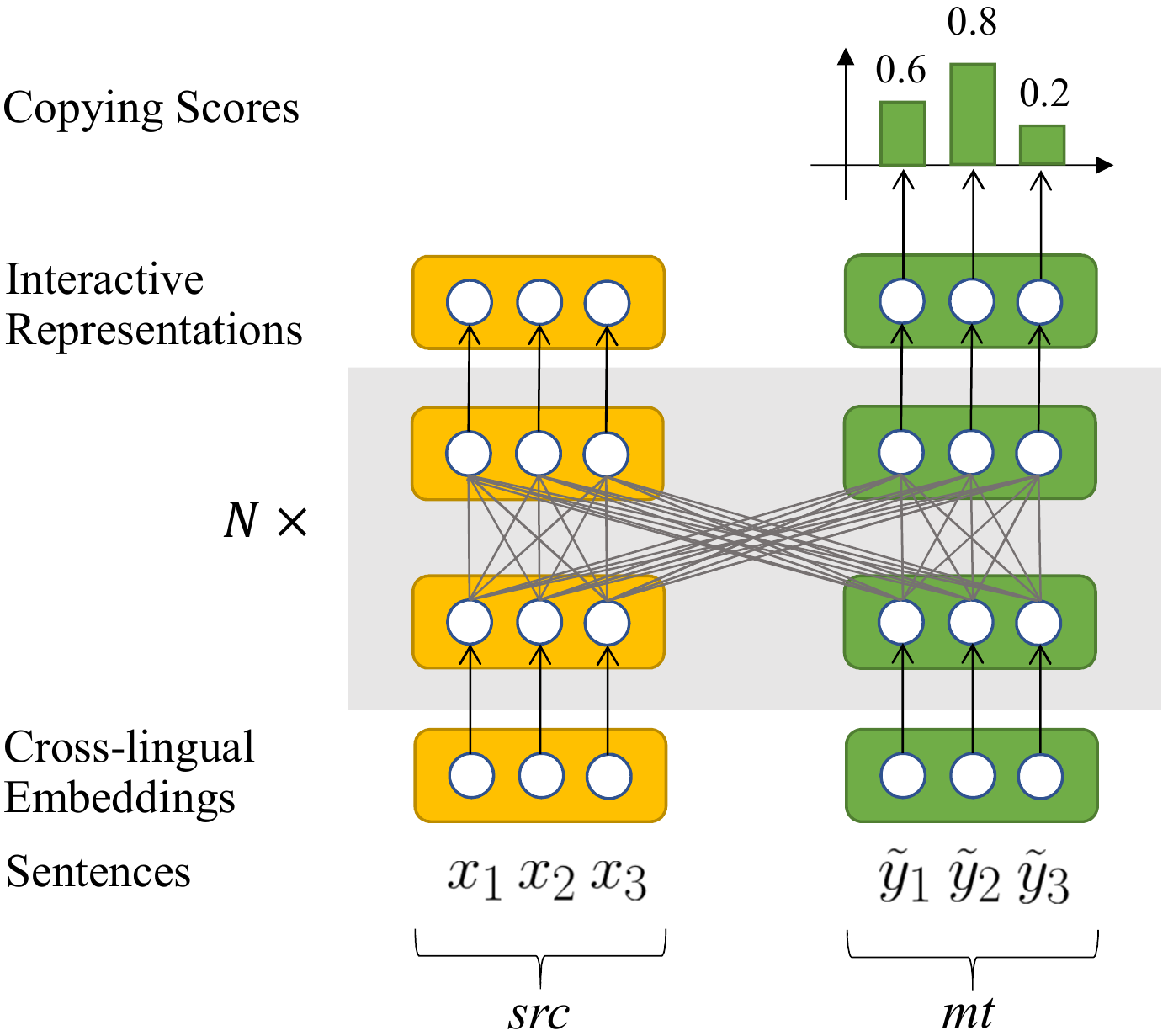}
  \caption{Interactive representation learning. Concatenated to serve as a single input, $src$ and $mt$ attend to each other during representation learning. Note that learnable weights of $src$ and $mt$ are shared. Interactive representation learning is used in both Predictor and Encoder. For example, copying scores are predicted based on the learned representations (see Eq. (\ref{eqn:s})). }
  \label{fig:inter_rep}
\end{figure}

\subsection{Interactive Representation Learning}
\label{sec:inter}
We propose an interactive representation learning method by making $src$ and $mt$ attend to each other. Following \citet{Lample2019CrosslingualLM} and \citet{He2018LayerWiseCB}, we concatenate $src$ and $mt$ in the dimension of sentence length with additional position and language embeddings:
\begin{alignat}{1}
\mathbf{X}_{i} &= \mathbf{E}^{\mathrm{token}}[x_i] + \mathbf{E}^{\mathrm{pos}}[i] + \mathbf{E}^{\mathrm{lang}}[0], \\
\mathbf{\tilde{Y}}_k &= \mathbf{E}^{\mathrm{token}}[\tilde{y}_k] + \mathbf{E}^{\mathrm{pos}}[k] + \mathbf{E}^{\mathrm{lang}}[1],
\end{alignat}
where $\mathbf{X}_{i}$ is the embedding of the $i$-th source word $x_i$, $\mathbf{\tilde{Y}}_k$ is the embedding of the $k$-th target word $\tilde{y}_k$, and $\mathbf{E}^{\mathrm{token}}$, $\mathbf{E}^{\mathrm{pos}}$, and $\mathbf{E}^{\mathrm{lang}}$ are the token, position and language embedding matrices. 

As shown in Figure \ref{fig:inter_rep}, the representation of $src$ and $mt$ can be learned jointly:
\begin{equation}
\begin{gathered}
\mathbf{H}^{\mathrm{inter}} = \mathrm{Encoder^{inter}}([\mathbf{X};\mathbf{\tilde{Y}}],\bm{\theta}), \label{eqn:H_inter}
\end{gathered}
\end{equation}
where $\mathrm{Encoder^{inter}}(\cdot)$ is the interactive Encoder and $[\mathbf{X};\mathbf{\tilde{Y}}]$ is the concatenation of $\mathbf{X}$ and $\mathbf{\tilde{Y}}$ in the dimension of sentence length.

As shown in Figure \ref{fig:architecture}(b), the multi-source Encoders are replaced by the interactive Encoder, which enables $src$ and $mt$ to attend to each other.\footnote{Note that the Predictor and Encoder do not share their weights because we found that sharing leads to degraded performance in experiments.} We expect that enabling the interactions between them can help to strengthen the ability of the model to find which words in $src$ is untranslated and which words in $mt$ is correct. Note that interactive representation learning is used both in Predictor and Encoder. In the following, we will describe how to predict which words in $mt$ should be copied based on these learned representations.  


\subsection{Predicting Words to be Copied}
\label{sec:pred}
Given $mt$ and $pe$, we can label each word in $mt$ as 0 or 1. We use 1 to denote that the word is to be copied (e.g. ``Ich'' and ``einen" in Figure \ref{fig:example}) and 0 not to be copied (e.g. ``esse" and ``Hamburger" in Figure \ref{fig:example}). It is possible to use the Longest Common Sequence (LCS) \cite{Wagner1974The} algorithm to obtain common sequences between $mt$ and $pe$. If the word in $mt$ also appears in the common sequences, it will be labeled 1; otherwise, it will be labeled 0. We denote these labels as $l_1 \dots l_K$.

We propose a Predictor module to predict words to be copied. As discrete labels are non-differentiable during training, the Predictor module outputs \textbf{copying scores} instead for the target words in $mt$:
\begin{alignat}{1}
\mathbf{s} &= \mathrm{sigmoid}([\mathbf{H}^{\mathrm{pred}}_{I+1};\cdots;\mathbf{H}^{\mathrm{pred}}_{I+K}]  \mathbf{W}_s), \label{eqn:s}
\end{alignat}
where $\mathbf{s} \in \mathbb{R}^{K \times 1}$ is a vector of copying scores corresponding to the $K$ words in $mt$, $\mathbf{H}^{\mathrm{pred}} \in \mathbb{R}^{(I+K) \times d}$ is the representation of $src$ and $mt$:
\begin{alignat}{1}
\mathbf{H}^{\mathrm{pred}} &= \mathrm{Predictor}([\mathbf{X};\mathbf{\tilde{Y}}]; \bm{\theta}),
\end{alignat}
and $\mathbf{W}_s \in \mathbb{R}^{d \times 1}$ is a weight matrix. Only the representation of $mt$ (i.e., $[\mathbf{H}^{\mathrm{pred}}_{I+1};\cdots;\mathbf{H}^{\mathrm{pred}}_{I+K}]$) is used for calculating copying scores. \footnote{It is also possible to predict which source words in $src$ should be used to generate non-copied words in $pe$ (e.g., ``Kuchen'' in Figure \ref{fig:example}). This can be done by generating explicit labels using bilingual word alignment. We leave this for future work.}


As shown in Figure \ref{fig:architecture}(b), copying scores can be incorporated into three parts of our model: the Encoder, the Decoder, and the CopyNet. Inspired by \citet{Yang2018ModelingLF}'s strategy to integrate localness to self-attention, we propose to incorporate copying scores into our model by modifying attention weights involved in the aforementioned three modules.


The original scaled dot-product attention \cite{Vaswani2017AttentionIA} is defined as
\begin{alignat}{1}
\mathrm{energy} &= \frac{\mathbf{q}\mathbf{K}^\top }{\sqrt{d}}, \\
\mathrm{Att}(\mathbf{q},\mathbf{K}) &= \mathrm{softmax}(\mathrm{energy}), 
\end{alignat}
where $\mathbf{q} \in \mathbb{R}^{1 \times d}$ is the query vector, $\mathbf{K} \in \mathbb{R}^{(I+K) \times d}$ is the key matrix, and $\mathrm{energy} \in \mathbb{R}^{1 \times (I+K)}$ is the ``energy" vector.

\begin{figure}[!t]
  \centering
  \includegraphics[scale=0.55]{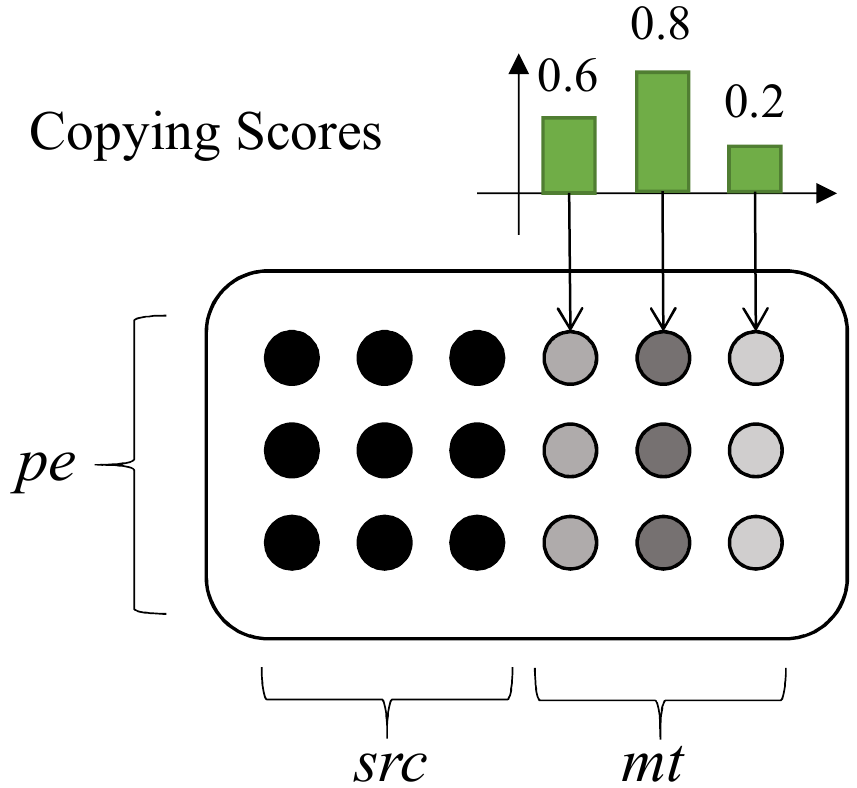}
  \caption{Copying scores as scaling masks. The copying scores are used to modify the attention layers of the Encoder, Decoder, and CopyNet.}
  \label{fig:att}
\end{figure}

As shown in Figure \ref{fig:att}, the copying scores can be used to form a scaling mask on the attention sub-layer:\footnote{Actually, we let ``energy" vector minus its minimum value to keep it non-negative.}
\begin{equation}
\mathrm{Att}(\mathbf{q},\mathbf{K})=\mathrm{softmax}\left(\mathrm{energy} \odot [\mathbf{m};\mathbf{s}]^\top\right),
\end{equation}
where $\mathbf{m} = \{ 1.0\}^{I}$ is a masking vector corresponding to $src$ and $\mathbf{s} \in \mathbb{R}^{K \times 1}$ is the vector of copying scores calculated by Eq. (\ref{eqn:s}) corresponding to $mt$. Note that copying scores are used to only change the attention weights related to $mt$ while the $src$ part is unchanged.


\subsection{Training}
\label{sec:training}
The training objective of our approach $L_{\mathrm{all}}(\bm{\theta})$ consists of three parts:
\begin{equation}
\begin{aligned}
L_{\mathrm{all}}(\bm{\theta}) =&\  (1-\alpha) \times (L_{\mathrm{ape}}(\bm{\theta}) + \lambda \times L_{\mathrm{copy}}(\bm{\theta})) \\
&+ \alpha \times L_{\mathrm{pred}}(\bm{\theta}), \label{eq:loss_all}
\end{aligned}
\end{equation}
where $\alpha$ and $\lambda$ are hyper-parameters.

The first part is the original log-likelihood loss function of APE: \footnote{For simplicity, we use sentence-level loss functions here. In practice, we use corpus-level loss functions.}
\begin{equation}
\begin{aligned}
L_{\mathrm{ape}}(\bm{\theta}) = -\log P(\mathbf{y}|\mathbf{x}, \mathbf{\tilde{y}}; \bm{\theta}).
\end{aligned}
\end{equation}

The second part is related to the CopyNet:
\begin{alignat}{1}
L_{\mathrm{copy}}(\bm{\theta}) &= \frac{1}{K}\sum_{k=1}^K (l_k-c_k)^2,
\end{alignat}
where $l_k$ is the ground-truth label (see Section \ref{sec:pred}) and $c_k$ is a quantity that measures how likely the $k$-th word in $mt$ to be copied by CopyNet:
\begin{alignat}{1}
c_k &= \sum_{j=1}^{J} \gamma_j \times P^{ \mathrm{copy}}(\tilde{y}_k).
\end{alignat}
Note that $\gamma \times P^{\mathrm{copy}}(y)$ is the term related to copying the target word $y$ in Eq. (\ref{eqn:copy_gate}).



The third part is a cross-entropy loss related to the Predictor:
\begin{equation}
\begin{aligned}
L_{\mathrm{pred}}(\bm{\theta}) = - \sum_{k=1}^K&\Big[l_k\mathrm{log}(s_k) \\
&+(1-l_k)\mathrm{log}(1-s_k)\Big],
\end{aligned}
\end{equation}
where $s_k$ is the copying score of the $k$-th word $\tilde{y}_k$ in $mt$.

Finally, we use an optimizer to find the model parameters that minimize the overall loss function:
\begin{eqnarray}
\hat{\bm{\theta}} = \argmin_{\bm{\theta}}\Big\{ L_{\mathrm{all}}(\bm{\theta}) \Big\}.
\end{eqnarray}

\section{Experiments}

\subsection{Setup}
\subsubsection*{Datasets}\label{sec:data}
We evaluated our approach on the WMT APE datasets, which often distinguish between two tasks: phrase-based statistical machine translation (i.e., PBSMT) and neural machine translation (i.e., NMT). All these APE datasets consist of English-German triplets containing source text ($src$), the translations ($mt$) from a ``black-box" MT system and the corresponding human-post-edits ($pe$). The statistics for the WMT APE datasets are shown in Table \ref{tab:data}. In addition to the official dataset, the organizers also recommend using additional datasets \cite{JunczysDowmunt2016LoglinearCO,Negri2018eSCAPEAL}. 

We used the WMT official dataset for the PBSMT task and the NMT task separately. The artificial training data \cite{JunczysDowmunt2016LoglinearCO} was also used for both tasks. More precisely, we used the concatenation of the		 official training data and the artificial-small data to learn a truecasing model \cite{Koehn2007MosesOS} and obtain sub-word units using byte-pair encoding (BPE) \cite{Sennrich2016NeuralMT} with 32k merges. Then, we applied truecasing and BPE to all datasets. We oversampled the official training data 20 times and concatenated them with both artificial-small and artificial-big datasets \cite{JunczysDowmunt2018MSUEdinST}. Finally, we obtained a dataset containing nearly 5M triplets for both tasks.  To test our approach on a larger PBSMT dataset, we used the eSCAPE synthetic dataset \cite{Negri2018eSCAPEAL}, which contains 7.2M sentences. By including the eSCAPE dataset, the training set is enlarged to nearly 12M sentences.

\begin{table}[!t]
    \begin{center}
        \begin{tabular}{|c|l|r|}
            \hline
            Category & Dataset & \# Sent. \\
            \hline
            \hline
            \multirow{4}{*}{PBSMT} 
             & training set & 23,000 \\
             & dev2016 & 1,000 \\
             & test2016 & 2,000 \\
             & test2017 & 2,000 \\
            \hline
            \multirow{2}{*}{NMT}
             & training set & 13,442 \\
             & dev2018 & 1,000 \\
            \hline
            \multirow{4}{*}{Additional}
             & artificial-small & 526,368 \\
             & artificial-big & 4,391,180 \\
             & eSCAPE-PBSMT & 7,258,533 \\
             & eSCAPE-NMT & 7,258,533 \\
            \hline
        \end{tabular}
        \caption{\label{tab:data} Statistics of the English-German datasets in the WMT APE task. Note that the NMT official data only contains training and development sets.}
    \end{center}
\end{table}

\subsubsection*{Hyper-Parameter Settings}
\label{sec:hyper-param}
For the original Transformer model, CopyNet and our approach, the hidden size was set to 512 and the filter size was set to 2,048. The number of individual attention heads was set to 8 for multi-head attention. We set $N = N_e = N_d = 6$, $N_p = 3$ and we tied all three $src, mt, pe$ embeddings for saving memory. The embeddings and softmax weights were also tied. In training, we used Adam \cite{Kingma2015AdamAM} for optimization. Each mini-batch contains approximately 25K tokens. We used the learning rate decay policy described by \citep{Vaswani2017AttentionIA}. In decoding, the beam size was set to 4. We used the length penalty \cite{Wu2016GooglesNM}
and set the hyper-parameter to 1.0. The other hyper-parameter settings were the same as the Transformer model \cite{Vaswani2017AttentionIA}. We implemented our approach on top of the open-source toolkit THUMT \cite{Zhang2017THUMTAO}.\footnote{https://github.com/THUNLP-MT/THUMT}

\begin{table}[!t]
    \begin{center}
        \begin{tabular}{|cc|cc|}
            \hline
            $\alpha$ & $\lambda$ & TER$\downarrow$ & BLEU$\uparrow$  \\
            \hline 
            \hline 
            0.1 & 1.0 & 18.83 & 72.59 \\
            0.5 & 1.0 & 18.45 & 72.83 \\
            0.9 & 0.1 & 18.89 & 72.27 \\
            0.9 & 0.5 & 18.47 & 72.83 \\
            0.9 & 1.0 & \textbf{18.38} & \textbf{72.99} \\
            \hline
        \end{tabular}
        \caption{\label{tab:alphaAndlambda} Effect of $\alpha$ and $\lambda$. The TER and BLEU scores are calculated on the WMT 2016 APE official development set.}
    \end{center}
\end{table}

\begin{table*}[!t]
    \begin{center}
        \begin{tabular}{|l||cccc|cc|}
            \hline
            \multirow{2}{*}{System} & \multicolumn{2}{c}{TEST16} & \multicolumn{2}{c|}{TEST17} & \multicolumn{2}{c|}{TEST16+17} \\
            \cline{2-7}
             & TER$\downarrow$ & BLEU$\uparrow$ & TER$\downarrow$ & BLEU$\uparrow$ & TER$\downarrow$ & BLEU$\uparrow$  \\
            \hline
            \hline
            (1) \textsc{Original} & 24.76 & 62.11 & 24.48 & 62.49 & 24.62 & 62.30 \\
            \hline
            \hline
            \multicolumn{7}{|c|}{WMT 2017 official + artificial (5M)} \\
            \hline
            (2) \textsc{Npi-Ape} & 22.07 & 66.67 & 22.58 & 65.52 & -- & -- \\
            \hline   
            (3) \textsc{Postech} & 19.14 & 70.98 & 19.26 & 70.50 & -- & -- \\ 
            (4) \textsc{Fbk} & 18.79 & 71.48 & 19.54 & 70.09 & -- & -- \\
            (5) $\textsc{Ms\_Uedin}_{\text{ensemble}}$ & 18.86 & 71.04 & 19.03 & 70.46 & -- & -- \\
            \hline
            (6) \textsc{CopyNet} & 18.91 & 71.64 & 19.47 & 70.75 & 19.19 & 71.19 \\
            \hline
            (7) Ours & 18.39 & 72.50 & 18.81 & 71.35 & 18.60 & 71.92 \\
            (8) $\text{Ours}_{\text{ensemble}}$ & \textbf{17.77} & \textbf{73.19} & \textbf{18.41} & \textbf{72.09} & \textbf{18.09} & \textbf{72.62} \\
            \hline
            \hline
            \multicolumn{7}{|c|}{WMT 2017 official + artificial + eSCAPE (12M)} \\
            \hline
            (9) \textsc{Usaar\_Dfki} & -- & 68.52 & -- & 68.91 & -- & -- \\
            (10) $\textsc{Ms\_Uedin}_{\text{ensemble}}$ & 17.34 & 73.43 & 17.47 & 72.84 & -- & -- \\
            \hline
            (11) \textsc{CopyNet} & 18.06 & 72.77 & 18.29 & 71.89 & 18.18 & 72.37  \\
            \hline
            (12) Ours & 17.45 & 73.51 & 17.77 & 72.98 & 17.61 & 73.24 \\
            (13) $\text{Ours}_{\text{ensemble}}$ & \textbf{17.06} & \textbf{74.00} & \textbf{17.37} & \textbf{73.26} & \textbf{17.22} & \textbf{73.62} \\
            \hline
        \end{tabular}
        \caption{\label{tab:pbsmt} Results on the English-German PBSMT sub-task. ``TEST16+17" is the concatenation of ``TEST16" and ``TEST17". $\textsc{Ms\_Uedin}_{\text{ensemble}}$ and $\text{Ours}_{\text{ensemble}}$ used ensembles of four models.}
    \end{center}
\end{table*}

\subsubsection*{Evaluation Metrics}
\label{sec:eval}
We used the same evaluation metrics as the official WMT APE task \cite{Chatterjee2018FindingsOT}: case-sensitive BLEU and TER. BLEU is computed by {\em multi-bleu.perl} \cite{Koehn2007MosesOS}. TER is calculated using TERcom.\footnote{http://www.cs.umd.edu/˜snover/tercom/}


\subsubsection*{Baselines}
We compared our approach with the following seven baselines: 
\begin{enumerate}
    \item \textsc{Original}: the original $mt$ without any post-editing.
    \item \textsc{CopyNet} \cite{Gu2016IncorporatingCM, Zhao2019ImprovingGE}: the multi-source Transformer equipped with CopyNet (see Figure \ref{fig:architecture}(a)). 
    \item \textsc{Npi-Ape} \cite{vu2018automatic}: a neural programmer-interpreter approach. 
    \item \textsc{Ms\_Uedin} \cite{JunczysDowmunt2018MSUEdinST}: a multi-source Transformer-based APE system that shares the encoders of $src$ and $mt$. It is the champion of the WMT 2018 APE shared task.
    \item \textsc{Usaar\_Dfki} \cite{Pal2018ATM}: a multi-source Transformer-based APE system with a joint encoder that attends over a combination of two encoded sequences. It is a participant of the WMT 2018 APE shared task.
    \item \textsc{Postech} \cite{Shin2018MultiencoderTN}: a multi-source Transformer-based APE system with two encoders. It is a participant of the WMT 2018 APE shared task.
    \item \textsc{Fbk} \cite{Tebbifakhr2018MultisourceTF}: a multi-source Transformer-based APE system with two encoders. It is a participant of the WMT 2018 APE shared task.
\end{enumerate}

We implemented \textsc{CopyNet} also on top of THUMT \cite{Zhang2017THUMTAO}. The results of all other baselines were taken from the corresponding original papers.

\subsection{Effect of Hyper-parameters}
We first investigated the effect of the hyper-parameters $\alpha$ and $\lambda$ in Eq. (\ref{eq:loss_all}). As shown in Table \ref{tab:alphaAndlambda}, using $\alpha=0.9$ and $\lambda=1.0$ achieves the best performance in terms of TER and BLEU on the WMT 2016 development set, suggesting that both the Predictor and CopyNet play important roles in our approach. Therefore, we set $\alpha=0.9$ and $\lambda=1.0$ in the following experiments.



\subsection{Main Results}

\subsubsection*{Results on the PBSMT Sub-task}

Table \ref{tab:pbsmt} shows the results of the PBSMT sub-task. We used the development set of the WMT 2016 APE PBSMT sub-task for model selection for our approach. 

Under the small-data training condition (i.e., 5M), our single model (i.e., System 7) outperforms all single-model baselines on all test sets. The superiority over \textsc{CopyNet} (i.e., System 6) suggests that interactive representation learning and incorporating copying scores are effective in improving APE. Our approach that uses the ensemble of four models (i.e., System 8) also improves over the best published result (i.e., System 5).

Under the large-data training condition (i.e., 12M), we find that our single (i.e., System 12) and ensemble (i.e., System 13) models still outperform the best single (i.e., System 11) and ensemble (i.e., System 10) models of baselines.


\begin{table}[!t]
    \begin{center}
        \begin{tabular}{|l|cc|}
            \hline
            System & TER$\downarrow$ & BLEU$\uparrow$ \\
            \hline
            \hline
            \textsc{Original} & 15.08 & 76.76 \\
            \hline
            \textsc{Postech} & 14.94 & 77.26 \\
            \hline
            \textsc{CopyNet}  & 15.12 & 77.05 \\
            \hline
            Ours & \textbf{14.88} & \textbf{77.40} \\
            \hline
        \end{tabular}
        \caption{\label{tab:nmt} Experiments on the WMT 2018 English-German NMT sub-task. The TER and BLEU scores are calculated on the WMT 2018 APE NMT sub-task development set.}
    \end{center}
\end{table}

\begin{table*}[!t]
    \begin{center}
        \begin{tabular}{|c|cccc||cc|cc|}
            \hline
            \multirow{2}{*}{ID} & \multicolumn{4}{c||}{Module} & \multicolumn{2}{c|}{DEV16} & \multicolumn{2}{c|}{TEST16+17} \\
            \cline{2-9}
             & Interactive & Predictor & CopyNet & Joint Training & TER$\downarrow$ & BLEU$\uparrow$ & TER$\downarrow$ & BLEU$\uparrow$  \\
            \hline
            \hline
            1 & $\surd$ & $\times$ & $\times$ & $\times$ & 18.74 & 72.21 & 19.11 & 71.03 \\
            2 & $\times$ & $\times$ & $\surd$ & $\times$ & 19.33 & 71.86 & 19.19 & 71.19 \\
            3 & $\times$ & $\surd$ & $\surd$ & $\surd$ & 19.11 & 72.03 & 19.07 & 71.30 \\
            4 & $\surd$ & $\times$ & $\surd$ & $\times$ & 18.62 & 72.42 & 18.91 & 71.53 \\
            5 & $\surd$ & $\surd$ & $\times$ & $\times$ & 18.53 & 72.53 & 18.85 & 71.54 \\
            6 & $\surd$ & $\surd$ & $\surd$ & $\times$ & 18.44 & 72.96 & 18.77 & 71.75 \\
            \hline
            7 & $\surd$ & $\surd$ & $\surd$ & $\surd$ & \textbf{18.38} & \textbf{72.99} & \textbf{18.60} & \textbf{71.92} \\
            \hline
        \end{tabular}
        \caption{\label{tab:ablation} Ablation study. ``Interactive'' denotes interactive representation learning (Section \ref{sec:inter}), ``Predictor'' denotes detecting correct words (Section \ref{sec:pred}), ``CopyNet'' denotes the CopyNet used in our approach (Section \ref{sec:CopyNet}), and ``Joint Training'' denotes the combination of three loss functions (Section \ref{sec:training}).}
    \end{center}
\end{table*}

\subsubsection*{Results on the NMT Sub-task}
Table \ref{tab:nmt} shows the results on the NMT sub-task. Besides \textsc{Original} and \textsc{CopyNet}, we also compared our approach with \textsc{Postech} \cite{Shin2018MultiencoderTN}, which is the only participating system that released the results on the development set of the WMT 2018 NMT sub-task.\footnote{As \textsc{Postech} only used small data for training, the eSCAPE datasets were not used in this experiment for a fair comparison.} We find that our approach also outperforms all baselines.


\subsection{Ablation Study}
Table \ref{tab:ablation} shows the results of ablation study. It is clear that interactive representation learning plays a critical role since removing it impairs post-editing performance (line 3). As shown in line 4, the Predictor is also an essential part of our approach. CopyNet and joint training are also shown to be beneficial for improving APE (lines 5 and 6) but seem to have relatively smaller contributions than interactive representation learning and predicting words to be copied.

\subsection{Results on Prediction Accuracy}

The Predictor is an important module in our approach as it predicts which words in $mt$ should be copied. Given the ground-truth labels, it is easy to calculate {\em prediction accuracy} by casting the prediction as a binary classification problem. We find that the Predictor achieves a prediction accuracy of 85.09\% on the development set.


\begin{table}[!t]
    \begin{center}
        \begin{tabular}{|l|c|}
            \hline
            System & Accuracy  \\
            \hline
            \hline
            \textsc{Ms\_Uedin} & 64.63 \\ 
            \textsc{CopyNet} & 64.72 \\ 
            \hline
            Ours & \textbf{65.61} \\ 
            \hline
        \end{tabular}
        \caption{\label{tab:acc} Comparison of copying accuracies.}
    \end{center}
\end{table}

\begin{figure*}[!t]
  \centering
  \includegraphics[scale=0.45]{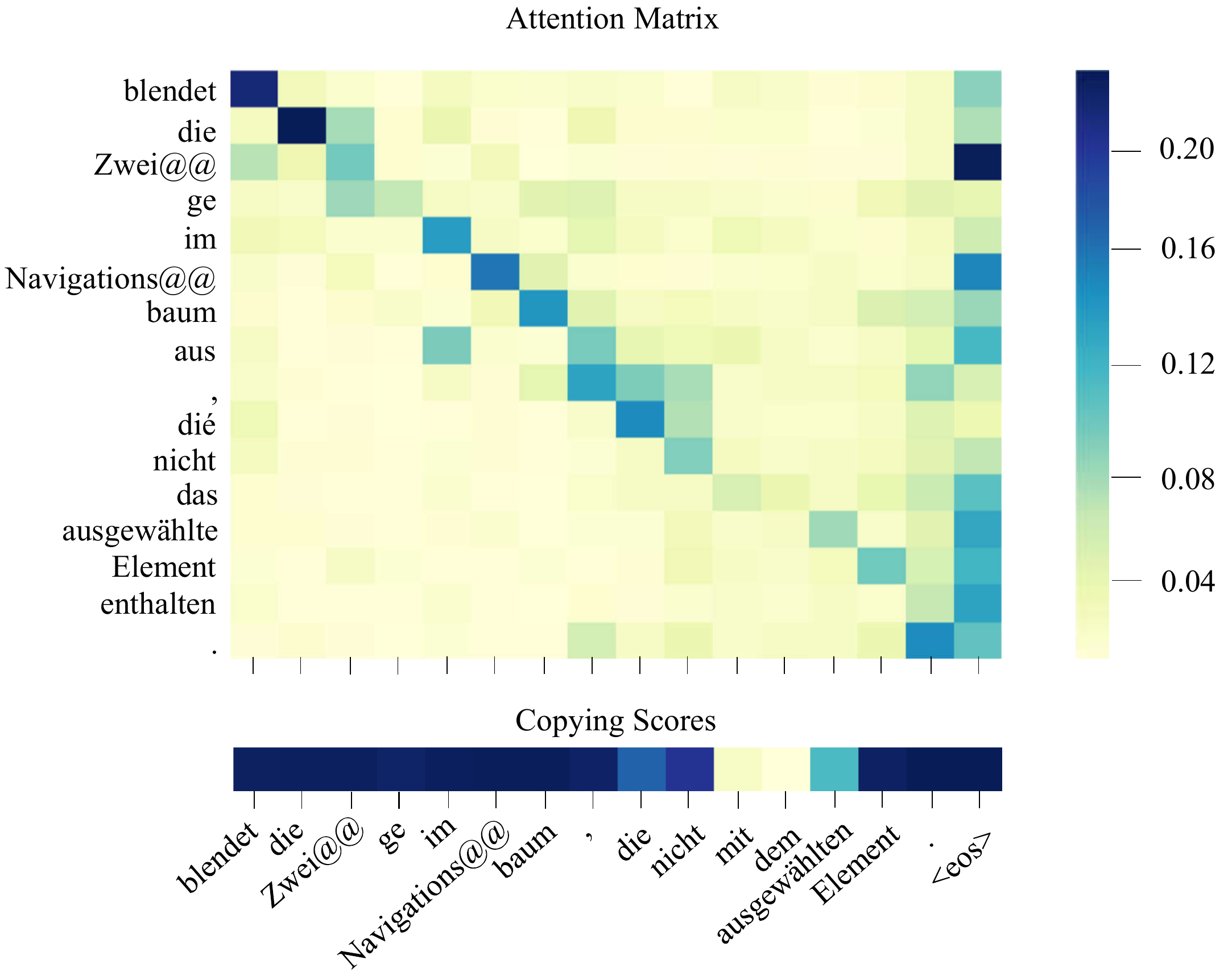}
  \caption{Example of the heatmap of attention and copying scores. The x-axis is $mt$ and the y-axis is $pe$. The Predictor successfully detects the incorrect word ``mit" and ``dem" and gives low copying scores to these words and then decrease the importance of them in attention. Other words in $mt$ are correctly copied to $pe$.}
  \label{fig:heatmap}
\end{figure*}

\subsection{Comparison of Copying Accuracies}

A target word $y$ in a machine translation $\tilde{\mathbf{y}}$ is called to be {\em correctly} copied to an automatic edited translation $\hat{\mathbf{y}}$ if and only if the positions where $y$ occurs in $\hat{\mathbf{y}}$ and $\mathbf{y}$ (i.e., the ground-truth edited translation) are identical. Therefore, it is easy to define copying accuracy to measure how well the copying mechanism works.

Table \ref{tab:acc} shows the comparison of copying accuracies between \textsc{Ms\_Uedin}, \textsc{CopyNet}, and our approach. We find that our approach outperforms the two baselines. However, the copying accuracy of our approach is almost 20\% lower than the prediction accuracy (i.e., 65.61\% vs. 85.09\%), indicating that it is much more challenging to place the copied words in correct positions.


\subsection{Visualization}
Figure \ref{fig:heatmap} gives an example that illustrates how copying scores influence attention. It shows the heatmap of the Enc-Dec-Attention, which averages over 8 different heads. Only the attention weights beween $pe$ and $mt$ are included. We take the second layer of Enc-Dec-Attention for example. The x-axis represents $mt$  and the y-axis represents $pe$. The darker the color, the higher the copying scores.  

We find that words ``mit" and ``dem" are identified by the Predictor. Accordingly, the attention weights corresponding to these words are decreased since the columns corresponding to these words have lighter color. As a result, all words in $mt$ other than ``mit" and ``dem" are copied to $pe$.


\section{Related Work}
\subsection{Multi-source Sequence-to-Sequence Learning}
Recently, multi-source Transformer-based APE systems \citep{JunczysDowmunt2018MSUEdinST, Pal2018ATM, Tebbifakhr2018MultisourceTF, Shin2018MultiencoderTN} have achieved the state-of-the-art results on the datasets of the WMT APE shared task. Our work differs from prior studies by enabling interactions between $src$ and $mt$ and explicitly detecting words to be copied.

\subsection{The Copying Mechanism}
\citet{Zhao2019ImprovingGE} apply CopyNet \cite{Gu2016IncorporatingCM} to grammar error correction. Their approach generates labels similar to ours, but only uses them to perform mutli-task learning. \citet{Libovick2016CUNISF} first introduce CopyNet to APE but do not provide a detailed description of their method and experimental results. We show that interactive representation learning and explicit indication of words are important for modeling copying in APE.

\subsection{Interactive Representation Learning}
\citet{Niehues2016PreTranslationFN} simply concatenate the output of the PBSMT system and the source sentence to serve as the input the NMT system without enabling multi-layer interactive learning. \citet{Lample2019CrosslingualLM} used cross-lingual setting to enable cross-lingual language model pre-training. We propose to let $src$ and $mt$ fully interact with each other to make it easier to decide which words in $mt$ should be copied.

\section{Conclusion}
We have presented a new method for modeling the copying mechanism for automatic post-editing. By making the source sentence and machine translation attend to each other, representations learned in such an interactive way help to identify whether a target word should be copied or be re-generated. We also find that explicitly predicting words to be copied is beneficial for improving the performance of post-editing. Experiments show that our approach achieves new state-of-the-art results on the WMT 2016 \& 2017 APE PBSMT sub-tasks.


\section*{Acknowledgments}
We thank all anonymous reviewers for their
valuable comments. This work is supported
by the National Key R\&D Program of China
(No. 2017YFB0202204), National Natural Science Foundation
of China (No. 61761166008, No.
61432013), Beijing Advanced Innovation Center
for Language Resources (No. TYR17002), and
the NExT++ project supported by the National Research Foundation,
Prime Ministers Office, Singapore under its IRC@Singapore Funding Initiative. This research is also supported by Sogou Inc.

\bibliography{emnlp-ijcnlp-2019}

\begin{thebibliography}{33}
\expandafter\ifx\csname natexlab\endcsname\relax\def\natexlab#1{#1}\fi

\bibitem[{B{\'e}chara et~al.(2011)B{\'e}chara, Ma, and van
  Genabith}]{Bchara2011StatisticalPF}
Hanna B{\'e}chara, Yanjun Ma, and Josef van Genabith. 2011.
\newblock Statistical post-editing for a statistical mt system.
\newblock In \emph{MT Summit}.

\bibitem[{Chatterjee et~al.(2017)Chatterjee, Farajian, Negri, Turchi,
  Srivastava, and Pal}]{Chatterjee2017MultisourceNA}
Rajen Chatterjee, M.~Amin Farajian, Matteo Negri, Marco Turchi, Ankit
  Srivastava, and Santanu Pal. 2017.
\newblock Multi-source neural automatic post-editing: Fbk's participation in
  the wmt 2017 ape shared task.
\newblock In \emph{Proceedings of WMT}.

\bibitem[{Chatterjee et~al.(2018)Chatterjee, Negri, Rubino, and
  Turchi}]{Chatterjee2018FindingsOT}
Rajen Chatterjee, Matteo Negri, Rapha{\"e}l Rubino, and Marco Turchi. 2018.
\newblock Findings of the wmt 2018 shared task on automatic post-editing.
\newblock In \emph{Proceedings of WMT}.

\bibitem[{Chatterjee et~al.(2015)Chatterjee, Weller, Negri, and
  Turchi}]{Chatterjee2015ExploringTP}
Rajen Chatterjee, Marion Weller, Matteo Negri, and Marco Turchi. 2015.
\newblock Exploring the planet of the apes: a comparative study of
  state-of-the-art methods for mt automatic post-editing.
\newblock In \emph{Proceedings of ACL}.

\bibitem[{Gu et~al.(2016)Gu, Lu, Li, and Li}]{Gu2016IncorporatingCM}
Jiatao Gu, Zhengdong Lu, Hang Li, and Victor O.~K. Li. 2016.
\newblock Incorporating copying mechanism in sequence-to-sequence learning.
\newblock In \emph{Proceedings of ACL}.

\bibitem[{He et~al.(2018)He, Tan, Xia, He, Qin, Chen, and
  Liu}]{He2018LayerWiseCB}
Tianyu He, Xu~Tan, Yingce Xia, Di~He, Tao Qin, Zhibo Chen, and T.~M. Liu. 2018.
\newblock Layer-wise coordination between encoder and decoder for neural
  machine translation.
\newblock In \emph{Proceedings of NeurIPS}.

\bibitem[{Junczys-Dowmunt and
  Grundkiewicz(2016)}]{JunczysDowmunt2016LoglinearCO}
Marcin Junczys-Dowmunt and Roman Grundkiewicz. 2016.
\newblock Log-linear combinations of monolingual and bilingual neural machine
  translation models for automatic post-editing.
\newblock In \emph{Proceedings of WMT}.

\bibitem[{Junczys-Dowmunt and Grundkiewicz(2017)}]{JunczysDowmunt2017AnEO}
Marcin Junczys-Dowmunt and Roman Grundkiewicz. 2017.
\newblock An exploration of neural sequence-to-sequence architectures for
  automatic post-editing.
\newblock In \emph{Proceedings of IJCNLP}.

\bibitem[{Junczys-Dowmunt and Grundkiewicz(2018)}]{JunczysDowmunt2018MSUEdinST}
Marcin Junczys-Dowmunt and Roman Grundkiewicz. 2018.
\newblock Ms-uedin submission to the wmt2018 ape shared task: Dual-source
  transformer for automatic post-editing.
\newblock In \emph{Proceedings of WMT}.

\bibitem[{Kingma and Ba(2014)}]{Kingma2015AdamAM}
Diederik~P. Kingma and Jimmy Ba. 2014.
\newblock Adam: A method for stochastic optimization.
\newblock In \emph{Proceedings of ICLR}.

\bibitem[{Knight and Chander(1994)}]{Knight1994AutomatedPO}
Kevin Knight and Ishwar Chander. 1994.
\newblock Automated postediting of documents.
\newblock In \emph{Proceedings of AAAI}.

\bibitem[{Koehn et~al.(2007)Koehn, Hoang, Birch, Callison-Burch, Federico,
  Bertoldi, Cowan, Shen, Moran, Zens, Dyer, Bojar, Constantin, and
  Herbst}]{Koehn2007MosesOS}
Philipp Koehn, Hieu Hoang, Alexandra Birch, Chris Callison-Burch, Marcello
  Federico, Nicola Bertoldi, Brooke Cowan, Wade Shen, Christine Moran, Richard
  Zens, Chris Dyer, Ondrej Bojar, Alexandra Constantin, and Evan Herbst. 2007.
\newblock Moses: Open source toolkit for statistical machine translation.
\newblock In \emph{Proceedings of ACL}.

\bibitem[{Lample and Conneau(2019)}]{Lample2019CrosslingualLM}
Guillaume Lample and Alexis Conneau. 2019.
\newblock Cross-lingual language model pretraining.
\newblock \emph{arXiv preprint arXiv:1901.07291}.

\bibitem[{Libovick{\'y} et~al.(2016)Libovick{\'y}, Helcl, Tlust{\'y}, Pecina,
  and Bojar}]{Libovick2016CUNISF}
Jindrich Libovick{\'y}, Jindrich Helcl, Marek Tlust{\'y}, Pavel Pecina, and
  Ondrej Bojar. 2016.
\newblock Cuni system for wmt16 automatic post-editing and multimodal
  translation tasks.
\newblock In \emph{Proceedings of WMT}.

\bibitem[{McKeown et~al.(2012)McKeown, Parton, Habash, Iglesias, and
  de~Gispert}]{McKeown2012CanAP}
Kathleen McKeown, Kristen Parton, Nizar Habash, Gonzalo Iglesias, and Adri{\`a}
  de~Gispert. 2012.
\newblock Can automatic post-editing make mt more meaningful?
\newblock In \emph{Proceedings of EAMT}.

\bibitem[{Negri et~al.(2014)Negri, Turchi, Chatterjee, and
  Bertoldi}]{Negri2018eSCAPEAL}
Matteo Negri, Marco Turchi, Rajen Chatterjee, and Nicola Bertoldi. 2014.
\newblock escape: a large-scale synthetic corpus for automatic post-editing.
\newblock In \emph{Proceedings of LREC}.

\bibitem[{Niehues et~al.(2016)Niehues, Cho, Ha, and
  Waibel}]{Niehues2016PreTranslationFN}
Jan Niehues, Eunah Cho, Thanh-Le Ha, and Alexander~H. Waibel. 2016.
\newblock Pre-translation for neural machine translation.
\newblock In \emph{Proceedings of COLING}.

\bibitem[{Pal et~al.(2018)Pal, Herbig, Kr{\"u}ger, and van
  Genabith}]{Pal2018ATM}
Santanu Pal, Nico Herbig, Antonio Kr{\"u}ger, and Josef van Genabith. 2018.
\newblock A transformer-based multi-source automatic post-editing system.
\newblock In \emph{Proceedings of WMT}.

\bibitem[{See et~al.(2017)See, Liu, and Manning}]{See2017GetTT}
Abigail See, Peter~J. Liu, and Christopher~D. Manning. 2017.
\newblock Get to the point: Summarization with pointer-generator networks.
\newblock In \emph{Proceedings of ACL}.

\bibitem[{Sennrich et~al.(2015)Sennrich, Haddow, and
  Birch}]{Sennrich2016NeuralMT}
Rico Sennrich, Barry Haddow, and Alexandra Birch. 2015.
\newblock Neural machine translation of rare words with subword units.
\newblock In \emph{Proceedings of ACL}.

\bibitem[{Shin and Lee(2018)}]{Shin2018MultiencoderTN}
Jaehun Shin and Jong-Hyeok Lee. 2018.
\newblock Multi-encoder transformer network for automatic post-editing.
\newblock In \emph{Proceedings of WMT}.

\bibitem[{Simard et~al.(2007)Simard, Goutte, and
  Isabelle}]{Simard2007StatisticalPP}
Michel Simard, Cyril Goutte, and Pierre Isabelle. 2007.
\newblock Statistical phrase-based post-editing.
\newblock In \emph{Proceedings of HLT-NAACL}.

\bibitem[{Tebbifakhr et~al.(2018)Tebbifakhr, Agrawal, Negri, and
  Turchi}]{Tebbifakhr2018MultisourceTF}
Amirhossein Tebbifakhr, Ruchit Agrawal, Matteo Negri, and Marco Turchi. 2018.
\newblock Multi-source transformer for automatic post-editing.
\newblock In \emph{Proceedings of CLiC-it}.

\bibitem[{Unanue et~al.(2018)Unanue, Borzeshi, and Piccardi}]{Unanue2018ASA}
Inigo~Jauregi Unanue, Ehsan~Zare Borzeshi, and Massimo Piccardi. 2018.
\newblock A shared attention mechanism for interpretation of neural automatic
  post-editing systems.
\newblock In \emph{Proceedings of NMT@ACL}.

\bibitem[{Varis and Bojar(2017)}]{Varis2017CUNISF}
Dusan Varis and Ondrej Bojar. 2017.
\newblock Cuni system for wmt17 automatic post-editing task.
\newblock In \emph{Proceedings of WMT}.

\bibitem[{Vaswani et~al.(2017)Vaswani, Shazeer, Parmar, Uszkoreit, Jones,
  Gomez, Kaiser, and Polosukhin}]{Vaswani2017AttentionIA}
Ashish Vaswani, Noam Shazeer, Niki Parmar, Jakob Uszkoreit, Llion Jones,
  Aidan~N. Gomez, Lukasz Kaiser, and Illia Polosukhin. 2017.
\newblock Attention is all you need.
\newblock In \emph{Proceedings of NeurIPS}.

\bibitem[{Vu and Haffari(2018)}]{vu2018automatic}
Thuy-Trang Vu and Gholamreza Haffari. 2018.
\newblock Automatic post-editing of machine translation: A neural
  programmer-interpreter approach.
\newblock In \emph{Proceedings of EMNLP}.

\bibitem[{Wagner and Fischer(1974)}]{Wagner1974The}
Robert~A. Wagner and Michael~J. Fischer. 1974.
\newblock The string-to-string correction problem.
\newblock \emph{Journal of the ACM}, 21:168--173.

\bibitem[{Wu et~al.(2016)Wu, Schuster, Chen, Le, Norouzi, Macherey, Krikun,
  Cao, Gao, Macherey, Klingner, Shah, Johnson, Liu, Kaiser, Gouws, Kato, Kudo,
  Kazawa, Stevens, Kurian, Patil, Wang, Young, Smith, Riesa, Rudnick, Vinyals,
  Corrado, Hughes, and Dean}]{Wu2016GooglesNM}
Yonghui Wu, Mike Schuster, Zhifeng Chen, Quoc~V. Le, Mohammad Norouzi, Wolfgang
  Macherey, Maxim Krikun, Yuan Cao, Qin Gao, Klaus Macherey, Jeff Klingner,
  Apurva Shah, Melvin Johnson, Xiaobing Liu, Lukasz Kaiser, Stephan Gouws,
  Yoshikiyo Kato, Taku Kudo, Hideto Kazawa, Keith Stevens, George Kurian,
  Nishant Patil, Wei Wang, Cliff Young, Jason Smith, Jason Riesa, Alex Rudnick,
  Oriol Vinyals, Gregory~S. Corrado, Macduff Hughes, and Jeffrey Dean. 2016.
\newblock Google's neural machine translation system: Bridging the gap between
  human and machine translation.
\newblock \emph{arXiv preprint arXiv:1609.08144}.

\bibitem[{Yang et~al.(2018)Yang, Tu, Wong, Meng, Chao, and
  Zhang}]{Yang2018ModelingLF}
Baosong Yang, Zhaopeng Tu, Derek~F. Wong, Fandong Meng, Lidia~S. Chao, and Tong
  Zhang. 2018.
\newblock Modeling localness for self-attention networks.
\newblock In \emph{Proceedings of EMNLP}.

\bibitem[{Zhang et~al.(2017)Zhang, Ding, Shen, Cheng, Sun, Luan, and
  Liu}]{Zhang2017THUMTAO}
Jiacheng Zhang, Yanzhuo Ding, Shiqi Shen, Yong Cheng, Maosong Sun, Huan-Bo
  Luan, and Yang Liu. 2017.
\newblock Thumt: An open source toolkit for neural machine translation.
\newblock \emph{arXiv preprint arXiv:1706.06415}.

\bibitem[{Zhao et~al.(2019)Zhao, Wang, Shen, Jia, and
  Liu}]{Zhao2019ImprovingGE}
Wei~Ke Zhao, Liang Wang, Kewei Shen, Ruoyu Jia, and Jingming Liu. 2019.
\newblock Improving grammatical error correction via pre-training a
  copy-augmented architecture with unlabeled data.
\newblock In \emph{Proceedings of NAACL-HLT}.

\bibitem[{Zoph and Knight(2016)}]{Zoph2016MultiSourceNT}
Barret Zoph and Kevin Knight. 2016.
\newblock Multi-source neural translation.
\newblock In \emph{Proceedings of HLT-NAACL}.

\end{thebibliography}
\bibliographystyle{acl_natbib}

\end{document}